\documentclass[
]{ceurart}

\usepackage{listings}
\usepackage{graphicx}
\usepackage{url}
\usepackage{subfigure} 

\begin{document}


\conference{8th International Workshop on Computational Models of Narrative}

\title{Collaborative Storytelling and LLM: A Linguistic Analysis of Automatically-Generated Role-Playing Game Sessions}


\author[1]{Alessandro Maisto}[%
orcid=0000-0002-3854-0470,
email=amaisto@unisa.it
]
\cormark[1]
\address[1]{Università degli Studi di Salerno, Via Giovanni Paolo II, 132, 84084, Fisciano (SA), Italia}

\cortext[1]{Corresponding author.}

\begin{abstract}
 Role-playing games (RPG) are games in which players interact with one another to create narratives. The role of players in the RPG is largely based on the interaction between players and their characters. This emerging form of shared narrative, primarily oral, is receiving increasing attention. In particular, many authors investigated the use of an LLM as an actor in the game. In this paper, we aim to discover to what extent the language of Large Language Models (LLMs) exhibit oral or written features when asked to generate an RPG session without human interference. We will conduct a linguistic analysis of the lexical and syntactic features of the generated texts and compare the results with analyses of conversations, transcripts of human RPG sessions, and books. We found that LLMs exhibit a pattern that is distinct from all other text categories, including oral conversations, human RPG sessions and books. Our analysis has shown how training influences the way LLMs express themselves and provides important indications of the narrative capabilities of these tools.
\end{abstract}

\begin{keywords}
Role-playing Games \sep
Large Language Models \sep
Shared Narrative \sep
Linguistic Analysis
\end{keywords}

\maketitle

\section{Introduction}
Despite the introduction of writing and the so-called \textit{Gutenberg Galaxy}, which revolutionized human narrative, many kinds of stories remain characterized by orality. Children's stories, jokes, anecdotes, and other everyday narrations are often expressed orally due to their occasional and informal nature. However, a contemporary form of shared narrative, primarily oral, is receiving increasing attention: the Role-Playing Game (RPG).

RPG narratives have two interesting features \cite{tychsen2006role,cover2014creation}:
\begin{itemize}
    \item Collaborative Storytelling: the narrative is cooperative, with stories built by a group of people. One person acts as the "moderator" or Game Master (GM), a neutral narrator who guides the story, manages the setting, and facilitates interactions between the Player Characters (PCs). Each PC, portrayed by a player, represents an independent actor within the story.
    \item Primarily Oral Narrative: although the GM may use written materials (such as rulebooks or pre-planned adventures), and player characters have their abilities defined on character sheets, the core of the narrative unfolds orally. The GM describes the situations and impersonates Non-Player Characters (NPCs), while the players describe their PCs' actions and dialogue, interacting with each other and the environment to achieve their goals.
\end{itemize}

Recently, Large Language Models (LLMs) have become widespread in RPGs because of their ability to generate stories and portray characters as virtual dialogue agents. Indeed, LLMs have proven useful in supporting GMs not only by generating random events and stories but also by making decisions for NPCs independently of the GM, adding unpredictability to their stories.

However, while an informal oral record characterises RPG sessions, the language of LLMs is often formal and may exhibit features typically associated with written narrative.

The objective of this paper is to discover to what extent the LLM language exhibits oral or written features. We will quantitatively and qualitatively analyze a dataset of automatically generated RPG sessions in which four LLM instances interact to generate a completely new story. We will conduct a linguistic analysis of the lexical and syntactic features of the generated texts and compare the results with analyses of oral conversations, transcripts of human RPG sessions, and books.

The paper is structured as follows: in the second section, we will briefly present the related works and the features of spoken and written discourses. In Section three, we describe our experiment and its results. Finally, in section five, we will extract our conclusions.

\section{Related Works}
Artificial Intelligence (AI), and particularly LLMs have been applied to the domain of RPGs in many different ways. Early work explored AI's ability to play text-adventure games, with research focusing on generating interactive worlds and automated quest generation \cite{fan2020generating,ammanabrolu2019toward}. These systems aimed to create engaging and dynamic game experiences by leveraging AI's ability to understand and respond to player choices in a text-based environment. 
More recent research has shifted towards using LLMs as virtual GMs in more complex RPG systems like \textit{Dungeons \& Dragons} (D\&D).  These models can generate stories, portray NPCs, and even make independent decisions for NPCs to enhance unpredictability \cite{callison2022dungeons,zhu2023fireball}. Several studies have investigated the potential of AI to participate in collaborative storytelling.  \citet{branch2021collaborative} explored how human actors and AI narrators can co-create stories, while \citet{zhang2023building} focused on building cooperative embodied agents using LLMs. These works highlight the potential of AI not just to generate stories but to actively participate in their creation alongside human players.

Beyond RPGs, research in story generation has explored various techniques to create engaging and coherent narratives. \citet{louis2018deep} used deep learning to model character interactions in D\&D transcripts, while \citet{park2023generative} developed generative agents that simulate human behaviour in interactive narratives.  Furthermore, researchers have tried to use common-sense reasoning to guide automated story generation and improve narrative coherence \cite{peng2022inferring}. Despite the progress, challenges remain in developing AI systems that can truly capture the nuances of human storytelling and social interaction in RPGs.  One key challenge is ensuring that AI-generated language aligns with the informal oral register typical of RPG sessions because the output of LLMs often exhibits features associated with written text.

Differences between orality and writing have been widely studied since the last century. The fundamental differences between oral and written cultures have been investigated by \citet{ong1982orality}. The main interest of Ong was in primary oral cultures (i.e. groups of people that never known the writing). Although our interest lies not in oral cultures but solely in the differences between written and spoken narrative, some characteristics of orality highlighted by Ong may prove relevant. In particular, Ong defines oral thinking and expression as "additive rather than subordinate", with a prevalence of coordination mechanisms; as "aggregative rather than analytic", rich in epithets and parallel or opposite terms; "redundant"; "empathetic and participatory rather than objectively distanced"; "homeostatic", strongly anchored in the present; "situational rather than abstract", characterized by concreteness rather than abstraction.

Regarding the characteristics of spoken discourse, 
\citet{ochs1979planned} and 
\citet{Chafe+1984+1095+1102} identified a large list of features that in some cases correspond to the features of orality by Ong. He affirmed that written languages are characterized by a high degree of "integration" in opposition to the "fragmentation" of spoken languages. Instead, spoken languages are characterized by "involvement", in opposition to the "detachment" typical of written languages. Chafe's list of features directly refers to "integration" and "involvement". Nevertheless, \citet{tannen1982oral} remarks that those notions "operate on different levels and are not mutually exclusive". \citet{tannen1982oral} proposed a comparative analysis of spoken and written versions of narratives, demonstrating "(1) that features which have been identified as characterizing oral discourse are also found in written discourse, and (2) that the written short story combines syntactic complexity, expected in writing, with features that create involvement, expected in speaking".

\subsection{Spoken and Written Discourse Features}
As \citet{tannen1982oral} pointed out, written narratives can combine features of spoken discourse and formal writing, especially in short stories. A single discourse type can combine the features of "integration" with the features of "involvement". \citet{tannen1982oral} suggests that literary language, in general, "uses features of involvement to create the kind of knowing that \citet{havelock1963preface} shows to be basic to oral performance: subjective knowing, established through a sense of identification between audience and performer or audience and characters in the text."

\citet{tannen1982oral} also reports some recurring hypotheses regarding written and spoken discourse. \citet{olson1977utterance}, for example, affirmed that spoken discourse is context-bound, while written discourse is de-contextualized. Cohesion is lexically established in writing but relies on paralinguistic and non-verbal channels in speech. However, \citet{tannen1982oral} affirmed that the former characteristic is related to the genre of discourse (casual conversation versus expository prose), while the latter is "a necessary concomitant of spoken and written modes."

Keeping this assumption in mind, typical distinctions between informal speech and planned written discourse have been identified by \citet{ochs1979planned} and \citet{Chafe+1984+1095+1102}. Table \ref{tab:1} shows the respective features.

\begin{table*}[htb]
  \caption{Characteristics of Spoken and Written Discourses}
  \label{tab:1}
  \footnotesize
  \begin{tabular}{p{7cm}|p{7cm}}
    \toprule
   \textbf{ Ochs Spoken Feature} & \textbf{Ochs Written Features}\\
    \midrule
    Use of simple morphosyntactic structures & Complex morphosyntactic structures \\
    Cohesion expressed by the immediate context & Cohesion expressed by lexical devices \\
    Preference for deictic modifiers (e.g. \textit{this}, \textit{that}, \textit{here}, etc.) & Preference for definite and indefinite articles \\
    Avoidance of relative clauses & Use of Relative clauses\\
    Preponderance of repair mechanisms & Absence of repair mechanisms \\
    Use of parallelism (phonetic, lexical and syntactic repetitions) & Less use of parallelism \\
    Tendency to begin in past tense and switch to present & Tendency to use past tense \\
    \midrule
    \textbf{Chafe Spoken Features} & \textbf{Chafe Written Features}\\
    \midrule
    Concreteness and imageability through specific details & Nominalization \\
    Use of 1st person  & Use of participles \\
    Emphasis on people and relationships & Attributive adjectives \\
    Emphasis on actions and agents  & Conjoined phrases \\
    Direct quotation  & Sequences of prepositional phrases \\
    Reports of speaker's mental processes & Complement clauses \\
    Fuzziness & Relative clauses\\
    Emphatic particles &  \\
    Devices for monitoring communication channels & \\
  \bottomrule
\end{tabular}
\end{table*}

If we analyze the features shown in Table \ref{tab:1} in light of Ong's characteristics of orality, we can find some common elements. For example, Ong defined oral expression as additive rather than subordinate. Similarly, Ochs observed that spoken discourse relies on simple morphosyntactic structures acquired early in life, such as coordination, while Chafe highlighted the prevalence of subordinate clauses and prepositional phrases in written discourse.

Oral expression is aggregative. This means that, for example, a noun is often accompanied by the same adjective, resulting in lexical or syntactic parallelisms. It is also empathetic and participatory, focusing on the speaker (through the use of the first person) or on emphatic particles. Furthermore, it is homeostatic, tending towards the use of the present tense. Finally, oral expression is typically concrete.

Some of the features reported in Table \ref{tab:1} can be easily verified computationally, while others are difficult to formalize in a computer program. For example, "reports of speaker's mental processes" or "Devices for monitoring communication channels" can be automatically analyzed only if we rely on specific semantic databases (for the tag of concepts related to mental processes or actions, agents, etc.) or if we have transcriptions that report non-verbal behaviours. 

In the next section, we will associate the computationally verifiable features with standard automatic text analysis metrics to compare the language produced by a Large Language Model (LLM) with transcribed spoken discourse and written texts.

\section{Experiment}
To analyze the characteristics of the language produced by Large Language Models (LLMs), we analyzed a dataset of automatically generated texts and positioned them along a hypothetical continuum from spoken to written language. To do this, we compiled a diverse set of texts (sub-corpora), selecting eight different texts for each of the following categories:
\begin{itemize}
    \item \textbf{Conversations}: Transcribed spoken conversations from the British National Corpus (BNC)\cite{leech1992100};
    \item \textbf{Academic speeches}: Transcribed spoken academic discussions from the English as a Lingua Franca in the Academia (ELFA) corpus \cite{mauranen2006rich};
    \item \textbf{Human RPG sessions}: Transcriptions of D\&D sessions from the online web series Critical Role  (\url{https://criticalrole.fandom.com/});
    \item \textbf{Books}: Classic fantasy novels published between 1932 (Conan the Barbarian) and 1997 (Harry Potter and the Philosopher's Stone) (see Table \ref{tab:2}).
\end{itemize}

The sub-corpora were selected to facilitate a comparative analysis of RPG sessions with both spoken and written language corpora, specifically those representing conversational and narrative genres. The British National Corpus (BNC) was chosen due to its established reputation within the scientific community and its extensive collection of transcribed dialogues, providing a representative sample of everyday spoken English. In contrast, the ELFA corpus comprises academic speech, which exhibits distinct linguistic characteristics compared to the conversational language in the BNC. Notably including longer and more syntactically complex utterances. This corpus allows for an examination of spoken language used in a more formal and structured setting. Finally, the \textit{book} sub-corpus consists of classic works of fantasy literature. This selection ensures domain consistency with the RPG sessions, enabling a comparison of the LLM-generated language to established examples of narrative prose within the fantasy genre.

\begin{table*}
  \caption{Description of the Dataset}
  \label{tab:2}
  \footnotesize
  \begin{tabular}{ll|ll|ll|p{3cm}l}
    \toprule
   \textbf{conversation} & Tokens & \shortstack{\textbf{accademic} \\ \textbf{speech}} & Tokens & \textbf{human RPG} & Tokens & \textbf{book} & Tokens\\
    \midrule
    KBORE00A & 5341 & CDIS01A & 4309 & 1.2 & 28724 & Conan the Barbarian & 102731\\
    KBORE00B & 6937 & CDIS01B & 3311 & 1.10 & 42139 & The Hobbit & 95761\\
    KBORE00C & 4018 & CDIS01C & 3966 & 1.17 & 28922 & \shortstack{The Lion, the Witch \\ and the Wardrobe} & 37189\\
    KBORE00G & 3357 & CDIS01D & 5022 & 1.22 & 34720 & Elric of Melniboné 
    & 47744\\
    KBORE00J & 5841 & CDIS03A & 5309 & 2.30 & 23827 & The Elfstone of Shannara & 198338\\
    KBORE003 & 3222 & CDIS03B & 5404 & 2.37 & 36242 & Dragons of Autumn Twilight & 147942\\
    KBORE007 & 748 & USEMD130 & 10438 & 3.42 & 31935 & A Game of Thrones & 293418\\
    KBORE009 & 914 & USEMD150 & 2606 &  3.50 & 23361 & \shortstack{Harry Potter and \\ the Philosopher's Stone} & 78355\\
    \midrule
    Average & 3797.25 & Average & 5045.625 & Average & 31233.75 & Average & 125184.75 \\
    \bottomrule
\end{tabular}
\end{table*}

Table \ref{tab:2} reports the file codes from the BNC and ELFA corpora. The CR texts are named using the campaign number followed by the number of the recorded session (for example, file 2.37 corresponds to the transcript of session 37 from the second campaign). Critical Role transcripts have been used previously for research purposes, as demonstrated by \citet{rameshkumar2020storytelling}. The dataset was downloaded from \url{https://criticalrole.fandom.com/}, which also contains descriptions of the player characters and summaries of each session.

This paper aims to compare the documents in our corpus to a set of RPG sessions generated entirely by an LLM (\textbf{LLM RPG} sub-corpus). To do this, we generated four different instances of an LLM to automatically generate a conversation among them. To accelerate the computational process, we used the Groq API (\url{https://groq.com/}), which provides access to high-speed hardware that enables fast response times and supports numerous LLMs, including LLaMa3-8b-8192, which we used in our experiment.

\begin{figure}[htb]
    \centering
    \includegraphics[width=0.8\linewidth]{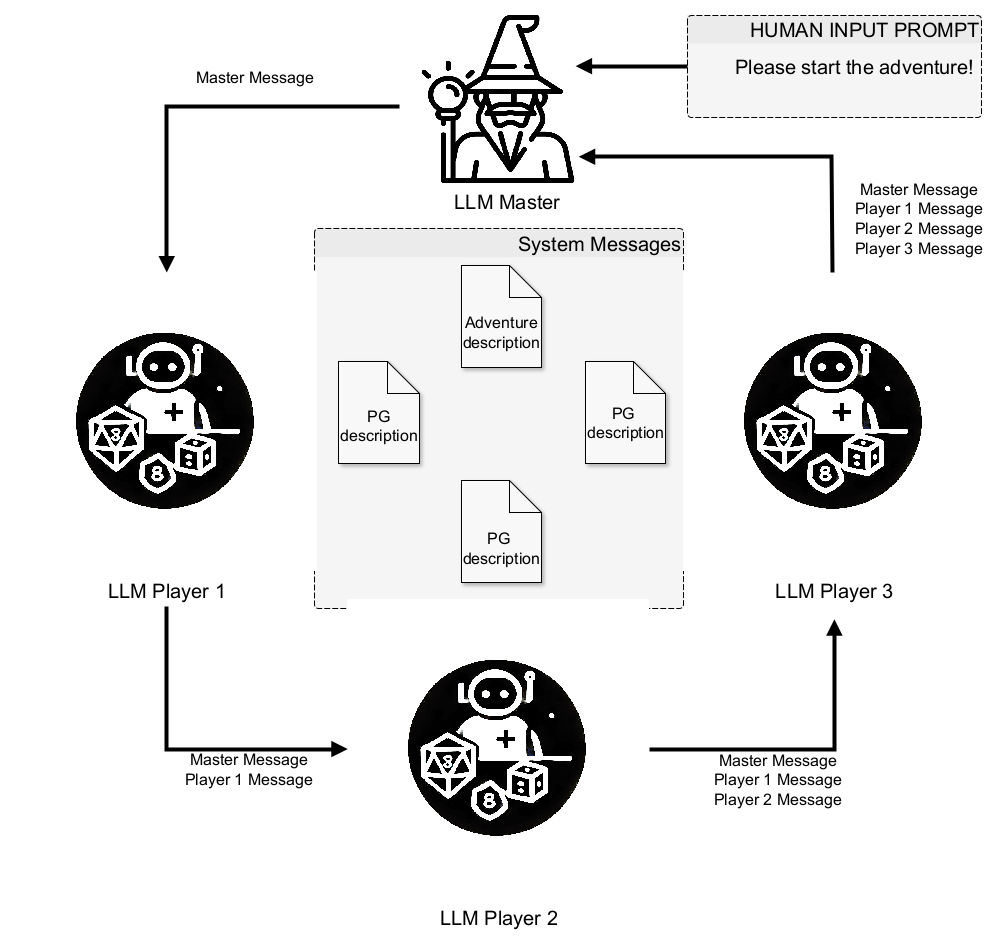}
    \caption{LLM Session Scheme}
    \label{fig:1}
\end{figure}

As shown in Figure \ref{fig:1}, we generated four LLM instances: one instance represents the Game Master (GM), and the remaining three represent players. Each instance was provided with a specific system message containing the adventure description for the GM and the character description for each player:

\begin{quote}
    \textbf{Master System Message}:\\
You are the Dungeon Master of a fantasy role-playing game. Your task is to guide a group of adventurers through a dungeon filled with monsters, traps, and treasure. The Adventure is ...\\
\\
\textbf{Player System Message}:\\
You are a Dungeons and Dragons player and take the role of ...

\end{quote}

Each instance was configured with a specific temperature and a maximum number of tokens for each response. Furthermore, the player LLMs were provided with descriptions of player characters from the Critical Role dataset, specifically Grog, Pike, and Vax. Table \ref{tab:3} summarizes the features of each instance created for the automatic generation of the RPG session.

\begin{table*}[htb]
\caption{Features of LLM's Instances in different sessions}
\centering
\footnotesize
\begin{tabular}{|l|c|c|c|c|c|c|}
\hline
\textbf{Instance} & \multicolumn{2}{c|}{\textbf{Session 1-2}} & \multicolumn{2}{c|}{\textbf{Session 3-4}} & \multicolumn{2}{c|}{\textbf{Session 5-8}} \\ \hline
 & \textbf{Temperature} & \textbf{Tokens} & \textbf{Temperature} & \textbf{Tokens} & \textbf{Temperature} & \textbf{Tokens} \\ \hline
\textbf{Master} & 0.3 & 200 & 0.5 & 200 & 1 & 200 \\ \hline
\textbf{Grog} & 0.4 & 100 & 0.7 & 100 & 0.7 & 100 \\ \hline
\textbf{Pike} & 0.3 & 100 & 0.5 & 100 & 0.5 & 100 \\ \hline
\textbf{Vax} & 0.3 & 50 & 0.4 & 50 & 0.4 & 50 \\ \hline
\end{tabular}
\label{tab:3}
\end{table*}

Descriptions of each character were manually extracted from the character information provided on the Critical Role wiki. An example of a character description is provided below:

\begin{quote}
You are a Dungeons and Dragons player and take the role of Grog, a goliath barbarian/fighter. Grog's low intelligence score has proven problematic for the group in the past. [...] Because of his low intelligence, Grog is illiterate and cannot read or write in any language. [...] Grog has shown himself to be rather impulsive, as seen often when he charges ahead of the group into an unknown area of a dungeon or map. [...] He was also the only chaotic-neutral group member, and he is often more aggressive and less merciful than some of the other characters. [...]\\    
\end{quote}

These sentences are an excerpt from the description provided in the system message to the LLaMa3 instance playing the role of Grog. For the adventure description provided to the Game Master instance, we used three different official D\&D adventure modules, automatically summarized by ChatGPT.  An extract of one of the adventure descriptions is provided below:

\begin{quote}
    The adventure takes place at Guido's Fort, an isolated frontier settlement on the banks of the River Shrill. Across the river lies The Hill, a mysterious and dangerous mountain full of deadly creatures, ancient ruins, and a multilevel dungeon.[...]\\ 
    \\
    Key Locations are:\\
    Guido's Fort: A frontier town where adventurers prepare. Inn rumors hint at various dangers, including ghouls, hobgoblins, and a dragon. Supplies and passage across the river are costly due to the area's danger.[...]\\
    \\
    Features include:\\
    Clearing with Killer Bees: A hive with honey acting as half-strength healing potions.\\
    Magical Berry Glade: Berries provide minor healing benefits but spoil quickly.[...]\\
    \\
    Key NPCs and Encounters are:\\
    The Sisters: Two mysterious, magical old women in a cottage who barter information or aid for gold or favors.[...]\\
    \\
    Starting the Adventure: Characters begin at Guido’s Fort and can ferry across the river to explore the hill.[...]
\end{quote}

The generated session begins with a manually introduced prompt: \textit{"Please start the adventure!"}. The Game Master then produces an answer, which serves as the introduction to the adventure. The first player to respond is always Grog, who receives the Game Master's answer as input. The second player, Pike, receives the concatenated responses of the Game Master and Grog as input.  Similarly, Vax receives the concatenated responses of the Game Master, Grog, and Pike. After the third player provides their response, the entire sequence of answers is passed back to the Game Master. The new response from the Game Master is then passed to Grog, and the session loop continues in this manner.

The loop is interrupted by the system when it reaches 200 interactions per player or if the API reaches the token limit.

Table \ref{tab:9} resumes the LLM sessions. Standard sessions were generated using the Dungeon Master Rulebook, with variations in parameters. Additionally, sessions were generated using two randomly selected adventure modules to examine differences in the narratives produced by the LLMs.

\begin{table*}[htb]
  \caption{Description of the Dataset}
  \label{tab:9}
  \footnotesize
  \begin{tabular}{lll}
    \toprule
   \textbf{Sessions} & \textbf{Adventure} & Tokens \\
    \midrule
    1 & Dungeon Master Rulebook - Your First Game, pp. 4-13 (1983) & 9259 \\
    2 & Dungeon Master Rulebook - Your First Game, pp. 4-13 (1983) & 16653 \\
    3 & Dungeon Master Rulebook - Your First Game, pp. 4-13 (1983) & 8324 \\
    4 & Horror on the Hill Adventure Module (1983) & 13640 \\
    5 & Horror on the Hill Adventure Module (1983) & 5360 \\
    6 & Dungeon Master Rulebook - Your First Game, pp. 4-13 (1983) & 13542 \\
    7 & Legions of Thyatis Adventure Module (1990) & 9184 \\
    8 & Legions of Thyatis Adventure Module (1990) & 9881 \\
    \midrule
     & Average Number of Tokens per Text & 10730.4 \\
    \bottomrule
\end{tabular}
\end{table*}

\subsection{Qualitative analysis}
A qualitative analysis of the LLM-generated data will yield a comprehensive understanding of the narrative mechanisms employed by these models.

Analysis of the generated sessions reveals a notable presence of Non-Player Characters (NPCs). These characters, typically controlled by the GM in traditional tabletop role-playing games, interact with PCs to guide the narrative trajectory.  By influencing the PCs' actions, the GM can effectively steer the narrative flow. 

While Prompt 3 ("Legions of Thyatis") explicitly defined the names and roles of certain NPCs, Prompts 1 ("Dungeon Master Rulebook Adventure") and 2 ("Horror on the hill") provided no such specifications.  In these instances, the LLM exhibited a tendency to repeat generic character designations across sessions. Notably, the name "Grimbold" was assigned to the innkeeper in both Session 1 and Session 2, as well as to the blacksmith in Sessions 3 and 6.  Furthermore, each Grimbold served a consistent narrative function: warning the PCs of impending danger. This repetition suggests the LLM may have been influenced by readily available online resources, as "Grimbold Ironfist" is a common NPC archetype in  Dungeons \& Dragons campaigns.

The antagonist varied across sessions, except in the final two, where the prompt explicitly designated this role.  Similarly, the LLM exhibited considerable variability in its generation of key objects, suggesting a degree of creativity in this domain. While an "(ancient) book" appeared in three sessions, a diverse array of other objects were also introduced, including a crystal, a map, a tapestry, a talking statue, a wooden box, a fragment of ancient knowledge, and a magic stone.

To analyze the narrative structure of the generated sessions, we employed Propp's framework of narrative functions \cite{propp2010morphology}, mapping each segment to a corresponding function\footnote{The 20 Propp's Functions retrieved in the LLM sessions are: Interdiction (k), Violation (q), Reconnaissance (v), Delivery (w), Trickery (j), Complicity (y), Villainy (X), Mediation (x) or Lack (X), Counteraction (W), Departure (↑), First Donor Function (D), Hero's Reaction (E), Receipt of Agent (Z), Struggle (L), Victory (V), Return (↓), Pursuit (P), Rescue (S), Difficult Task (C) and Solution (A). We also included in the representation the Initial Situation (i)}.

\begin{table*}[htb]
\caption{Propp's Function Schema for LLM Sessions}
\centering
\begin{tabular}{l|l}
\toprule
\textbf{Sessions} & \textbf{Narrative Functions}\\
\midrule
 1 & i k q ↑ j y w X D E Z X\\   
 2 & i C k q w ↑ L A ↓ P S\\ 
 3 & i k q ↑ L V v w Z k ↓ P S L V\\ 
 4 & i x k q ↑ C w W j L\\     
 5 & i Y W D L V Z k q v w\\    
 6 & i D E ↑ k j v\\       
 7 & i k q L P L V D P L S\\   
 8 & i D k x w L P L V P D j w L V P S\\
\bottomrule
\end{tabular}
\label{tab:6}
\end{table*}

Table \ref{tab:6} demonstrates a high degree of variability in the narrative structures generated, even among sessions initiated with identical prompts. While the violation of an interdiction (kq) followed by a departure (↑) appears consistently, the concluding sequences (epilogues) exhibit considerable variation.

Linguistic analysis reveals both similarities and noteworthy variations across the sessions. For instance, while the GM's opening remarks (incipit) within a given prompt share a common initial sentence structure, subsequent sentences diverge, as shown in table \ref{tab:7}.

\begin{table*}[htb]
    \caption{LLM GM's first sentences of different sessions}
    \footnotesize
    \centering
        \begin{tabular}{|p{5cm}|p{1.5cm}|p{8cm}|}
        \hline
        \multirow{3}{*}{\parbox{5cm}{\vspace{15mm} Welcome, adventurers, to the land of Eridoria, a realm of wonder and danger.}} & Session 1, Temp. 0.3 & Your journey begins in the small town of Willowdale, nestled at the foot of the mystical Dragon's Spine mountains \\ \cline{2-3} 
         & Session 3, Temp. 0.5 & Your journey begins in the bustling town of Willowdale, where rumors have been circulating about the long-abandoned castle of the powerful wizard, Malyster \\ \cline{2-3} 
        & Session 6, Temp. 1 & Your quest begins in the bustling town of Willowdale, nestled at the foot of the ancient castle ruins where the mighty wizard, Lyraea, once held court \\ \hline
        \hline
        \multirow{2}{*}{\parbox{5cm}{Welcome, adventurers, to Guido's Fort, a small but bustling frontier settlement on the banks of the River Shrill.}} & Session 4, Temp. 0.5 & The air is thick with the smell of smoke and sweat as the townsfolk go about their daily routines. \\ \cline{2-3} 
          & Session 5, Temp. 1 & The fort is a hub of activity, with travelers and traders arriving from all corners of the realm. \\ 
         \hline
         \hline
        Welcome to "Riot and Chaos at the Coliseum"! As you enter the Coliseum's barracks, you're greeted by the familiar sounds of clanging steel and the thrum of crowds. & Session 7, Temp. 1  & Yalag, the grizzled orcish gladiator trainer, looks up from his usual spot near the fire pit. His eyes narrow as he takes in the group of new faces. \\
        \hline
         Welcome, adventurers, to the city of Thyatis! You find yourselves in the bustling Coliseum, surrounded by the roar of the crowd and the glint of golden armor. & Session 8, Temp. 1 & You've been summoned by Yalag, the renowned gladiator trainer, who values your particular set of skills. \\ 
         \hline
    \end{tabular}
    \label{tab:7}
\end{table*}

Analysis of Table \ref{tab:7} suggests that the LLM's temperature setting influences the generated language, with lower values leading to more noticeable effects. For example, the description of Willowdale shifts from "small town" in Session 1 to "bustling town" in Sessions 2 and 3. However, the linguistic variations between Sessions 4 and 5 are less pronounced and warrant further investigation.

Generally, the language generated by the LLM GM with high-temperature settings exhibits similarities to the language employed by human GMs in the Critical Role transcripts. For instance, in File 2.30, the human GM recapitulates prior events and initiates the session with the following description: 
\begin{quote}
"As you guys come in from the soaking-wet rain that seemed to have been now pouring continuously through the city of Whitestone, you come to the quiet of your tavern, the sun beginning to find its way, setting across the evening sky"    
\end{quote}

Despite the use of the colloquial address term "guys" when addressing the players, this excerpt, characteristic of the human GM's style, also employs rich adjectival phrases, noun modifiers, and relative clauses. However, divergences between the LLM and human GM become more apparent in the dialogue generated for individual characters, as we will see below.

The responses generated for PCs present particularly compelling observations. The LLM appears to utilize language to reflect the distinct personalities defined in the initial prompt. Table \ref{tab:8} provides a comparative analysis of each PC's initial dialogue across varying temperature settings and adventures.

\begin{table*}[htb]
    \caption{LLM PCs first sentences of different sessions}
    \footnotesize
    \centering
        \begin{tabular}{|p{1cm}|p{1.5cm}|p{11cm}|}
        \hline
        \multirow{3}{*}{\parbox{2cm}{\vspace{20mm} Grog}} & Session 2, Temp. 0.4 & *grinning* Oooh, treasure! Grog love treasure! Grog ready to go! *picks up map and starts to study it, trying to make sense of the symbol and words* What mean "Beware the Guardian of the Gate"? Is it a big, scary monster? Grog want to fight it! *cracks knuckles* \\ \cline{2-3} 
         & Session 5, Temp. 0.7 &  *looks around at the crowd, squints at Oldrico* Grog no understand what Oldrico say. Grog want to know what treasure you talk about. Is it treasure of battle? Treasure of fight?\\ \cline{2-3}
          & Session 7, Temp. 0.7 &  *growling slightly, Grog steps forward, his massive frame towering over the others* "Grog no care about Senate or monsters. Grog here for... *pauses, thinking*... for FIGHTING!\\
        \hline
        \hline
        \multirow{3}{*}{\parbox{2cm}{\vspace{20mm} Pike}} & Session 2, Temp. 0.3 & *frowning* Ah, I don't know if I'm comfortable with this. I mean, we don't even know what kind of dangers we're facing. And what's with this "Guardian of the Gate" business? Sounds like a trap to me. *looks around at the others* Are we sure we're ready for this? *pauses, then looks back at the map*\\ \cline{2-3} 
         & Session 5, Temp. 0.5 & *steps forward, her eyes fixed intently on Oldrico* Ah, treasure, you say? *her voice is laced with a hint of excitement* Well, I'm always up for a good treasure hunt!\\ \cline{2-3}
         & Session 7, Temp. 0.5 & *smiling brightly, Pike's eyes sparkle with excitement* "Oooh, fighting, you say? Well, I'm all for that! I'm Pike, a cleric of the Everlight\\
         \hline
         \hline
        \multirow{3}{*}{\parbox{2cm}{\vspace{20mm} Vax}} &Session 2, Temp. 0.3 & *skeptical* I don't like it. We don't know what we're getting ourselves into. And what's with the cryptic map? It's like they're trying to lead us into a trap. *looks around at the others* I say we take a closer look around town, see if anyone's willing to talk about the castle and what's going on. Maybe we can gather some more information before we head out.\\ \cline{2-3} 
         & Session 5, Temp. 0.4 & *leaning against the wall of the inn, eyeing Oldrico with a mix of curiosity and skepticism* Ah, treasure, you say? *his voice is low and even* What kind of treasure are we talking about, Oldrico?\\ \cline{2-3}
          & Session 7, Temp. 0.4 & *leaning against the wall, eyeing the group with a mixture of curiosity and skepticism* "Fighting, huh? Sounds like a decent way to pass the time. I'm Vax, a half-elf rogue\\
         \hline
         \hline
    \end{tabular}
    \label{tab:8}
\end{table*}

The LLM-generated PC dialogues exhibit a distinct structure characterized by utterances heavily influenced by the provided character backgrounds. These utterances are interspersed with descriptions of non-verbal behaviors (e.g., movements, facial expressions, tone of voice), which tend to be stereotypical. For instance, Grog consistently employs simplistic, grammatically incorrect language, often utilizing capitalization for emphasis and using the third person to refer to himself ("Grog love treasure! Grog ready to go!"). Similarly, Vox is frequently portrayed as skeptical and cautious.

In contrast, human PC dialogues typically consist of concise, direct responses to the GM, often lacking detailed descriptions of non-verbal communication as in the following example, where Matt is the GM:

\begin{quote}
LIAM: I'm going to spend the first bit of time at home helping Pike, doing whatever she tells me to do, really, to help the people here.

MARISHA: Is there any deceased that need to be tended to?

MATT: Well, there are a number of deceased in the front courtyard, people that were frozen and shattered as part of Vorugal's attack, which at this point, most of them have been taken care of by the guards and given a proper burial on the outside of the Keep, a few yards away, towards the inside of the one of the hills. They haven't been left to thaw, if that was what you're worried about.

MARISHA: That's what I was worried about.

TALIESIN: Red meat. Red.

LAURA: Gross.

LIAM: Way to go dark. Way to go dark.

TALIESIN: I thought that's what we were doing today.
\end{quote} 

Overall, the LLM-generated dialogues tend to use descriptive language, employing a greater abundance of modifiers and adjectives compared to the human-generated dialogues. While the GM descriptions in both the LLM and human sessions share similarities, the PC dialogues in the human sessions are notably less complex, likely reflecting the spontaneous nature of spoken discourse.

This observation aligns with the findings of the quantitative analysis, which we will present in the next subsection.

\subsection{Automatic Feature extraction}
Some of the features proposed by Ochs and Chafe were formalized into a Java module created specifically for the project that automatically extracts them from the texts. We also included other characteristics used to evaluate the performance of the written narrative in the work of \citet{kormos2011task}. The measures considered are presented in Table \ref{tab:4}.

\begin{table*}[htb]
\caption{Summary of task performance measures used.}
\footnotesize
\centering
\footnotesize
\begin{tabular}{p{3cm}|p{3.5cm}|p{3.8cm}|p{2.7cm}}
\toprule
\textbf{Lexical Features} & \textbf{Syntactic Complexity} & \textbf{Analysis of Verbs} & \textbf{Frequency of Connectives}\\
\midrule
D-Value; & Sentence Length; & Present tense Ratio; & Additive;\\
Lexical Range; & Subordinate Ratio; & Past tense Ratio; & Causal;\\
Concreteness; & Relative Ratio; & Participle and Gerunds Ratio; & Temporal;\\
Ratio Deictics/Articles; & Distance from the ROOT; & 1st Person Pronoun Ratio & Logical\\
Lexical Repetitions; & Number of Noun Modifier; & & \\
Attributive Adjectives; & Average Dept of the Graph & \\
Emphatic particles & & & \\
\bottomrule
\end{tabular}
\label{tab:4}
\end{table*}

\subsubsection{Lexical Features}
We analyzed a variety of lexical features. First, we calculated the \textbf{D-value} \cite{malvern1997new}, a validated measure of lexical diversity \cite{mccarthy2010mtld}. The D-value is an alternative to the Type-Token Ratio and is less sensitive to text length. It works by calculating the probability that two randomly selected words are different.

We also calculated the \textbf{lexical range} (LR) as a measure of the lexical complexity of the texts. As suggested by \citet{kormos2011task}, we used the dictionaries of the Nation's Range program \cite{heatley2002range}, which calculates the percentage of words in a text that belong to the most frequent 1,000 (LR1) and 2,000 (LR2) words of "A General Service List of English Words" \cite{west1953general}, and the percentage of words belonging to "The Academic Word List" (LR3) by \citet{coxhead2000new}.

Although lexical diversity and range are not explicitly included in the list of features associated with orality, they can effectively capture differences between spoken and written texts. For instance, the use of formulas and repetitions in spoken discourse can affect the lexical diversity index and the lexical range of a text. To account for this, we also calculated the lexical repetitions found in the texts. \textbf{Lexical repetitions} were calculated by summing the uni-grams, bi-grams, and tri-grams repeated within a 2, 5, and 10-word window.

\textbf{Concreteness} ratings were based on the MRC Psycholinguistic Database \cite{coltheart1981mrc}, which provides concreteness values ranging from 100 to 700 for a large set of words. Concreteness is a specific feature of spoken discourse \cite{Chafe+1984+1095+1102}. To obtain a weighted concreteness index, we divided the sum of the concreteness values for all retrieved words in a text by the product of the number of tokens in the text and 100.

Frequencies of \textbf{deictics}, \textbf{attributive adjectives}, and \textbf{emphatic particles} were calculated by identifying the number of words in the text that belong to specific lists. These lists were compiled by collecting the most frequent elements of each category. We manually collected 69 deictics, 365 attributive adjectives, and 106 emphatic particles. As shown in Table \ref{tab:1}, emphatic particles and deictics are features of spoken discourse, while the presence of attributive adjectives characterizes written texts.

\subsubsection{Syntactic Complexity}
Syntactic complexity measures were calculated using the coreNLP Java package \cite{manning-EtAl:2014:P14-5}, which provides comprehensive syntactic tagging of texts. We calculated the \textbf{number of subordinate} and \textbf{relative clauses} using the dependency relation tags provided by coreNLP, corresponding to the Universal Dependency tags "csub," "ccomp," "advcl," "acl," "xcomp," and "parataxis." The tags "advcl:relcl" and "acl:relcl" specifically identify relative clauses. A large number of subordinate and relative clauses is characteristic of written text, as noted by \citet{Chafe+1984+1095+1102}, and is closely related to the use of complex morphosyntactic structures, which \citet{ochs1979planned} includes in their list of written features.

Syntactic complexity is also indicated by a high \textbf{average distance from the main element} of the sentence (the root node in the dependency graph, tagged as "ROOT") and by a high \textbf{graph depth}. The distance from the root is calculated as the average number of words separating the root from each word in the sentence. The depth of the graph is calculated as the maximum number of nodes separating the root from any other node.

Additionally, we calculated the number of \textbf{noun modifiers} ("nmod") as another measure of syntactic complexity.

\subsubsection{Analysis of Verbs}
As shown in Table \ref{tab:1}, a key difference between writing and speech lies in the use of \textbf{verb tenses}. We used the results of the syntactic parsing to calculate the percentage of verbs in the present tense (tagged as "VB," "VBZ," and "VBP"), past tense ("VBD"), and gerund or participle forms ("VBG," "VBN," and "VBG"). We also counted the number of \textbf{first-person pronouns} ("I," "me," "mine," "myself") as a feature of spoken discourse.

\subsubsection{Frequency of Connectives}
As suggested by \citet{kormos2011task}, we counted the number of connectives as a measure of text \textbf{cohesion}. Connectives are important devices for signaling relationships within a text \cite{halliday2014cohesion} and can provide insights into the textual quality of writing \cite{cameron1995text}. Following \citet{halliday2014cohesion}, we focused on four types of connectives: additive, causal, temporal, and logical. As noted by \citet{kormos2011task}, causal cohesion is particularly relevant when narrating events. We categorized connectives as either positive, extending the information provided in the text, or negative, restricting the information. We selected the 20 most frequent positive and negative connectives for each class and calculated the relative frequency of each class per 1,000 words of text. 

Next, we constructed a weighted cohesion score for each text. This score was calculated by summing the normalized frequencies of each connective class, with weights assigned to reflect the perceived contribution of each class to overall cohesion. Causal connectives received the highest weight (2), followed by logical and additive connectives (1.5) and temporal connectives (1).

To understand the distribution of cohesion scores across each corpus, we calculated the standard deviation. To further analyze the relationship between connective frequency and variability, we computed the logarithm of the standard deviation, using the weighted sum of frequencies as the base. This compressed the values and highlighted the differences among the analyzed classes.

\subsection{Results}
Starting with the lexical features, the differences between writing and speech are evident. Taking into account that the D-value is partially influenced by text length, we found a particularly high value in the traditional speech corpora (\textit{conversation}) and \textit{academic speech}, with the \textit{academic speech} texts reaching a value of 34.5. \textit{Books}, being the longest texts, achieved a value of 19.9. By comparing \textit{human RPG} and \textit{LLM RPG}, which are similar in length, we found a notable difference, with the human RPG reaching a D-value of 8.2 and the LLM ones 19.8.

Lexical Range (LR) values confirmed the differences between human and LLM RPG. While LR1 presented similar values across the different text categories, with higher values in the speech corpora, LR2 and LR3 showed the highest values with the book and LLM RPG corpora. Specifically, for LR2, the value for books was 0.060, for LLMs, it was 0.068, and for speech, it ranged from 0.037 (conversation) to 0.042 (human RPG). For LR3, the values increased from 0.013 for both speech corpora to 0.03 for human RPG and book, to 0.04 for LLM RPG. This indicates that LLMs use a wider range of vocabulary, including more academic and less frequent words, compared to humans when playing D\&D sessions.

Concreteness values confirmed the expectation that concreteness is a feature of spoken discourse. The \textit{conversation} sub-corpus presented the highest average concreteness value (12.82), followed by \textit{academic speech} (5.92). Interestingly, the value for \textit{human RPG} was 0.95, higher than the book sub-corpus (0.37) but lower than the value obtained for \textit{LLM RPG} (3.14). This suggests that while LLMs generate text with a degree of concreteness, they may overuse concrete terms compared to natural spoken language.

Regarding the presence of deictics, surprisingly, the sub-corpus with the highest ratio was \textit{LLM RPG} (0.41), followed by \textit{book} (0.37). speech corpora presented a lower ratio (0.32 for \textit{academic speech} and 0.30 for \textit{conversation}). \textit{Human RPG} had the lowest ratio, with a score of 0.16. These results do not align with the expectation that deictics primarily characterize spoken discourse. This discrepancy may be attributed to our list that includes not only personal, spatial, and temporal deictics but also social and discourse deictics, which may be more prevalent in written texts \cite{lyons1977semantics,levinson1983pragmatics,fillmore1997lectures}.

The analysis of emphatic particles yielded results consistent with the features presented in Table \ref{tab:1}. Speech presented values that were twice as high as those for books. In this case, the LLM-generated texts had a lower frequency of emphatic particles than books. Regarding the frequency of attributive adjectives, the differences among the different subsets were insignificant. More interesting was the frequency of repetitions. The values were particularly high for \textit{academic speech} and \textit{conversation}, but they were likely manually reduced in the \textit{human RPG} subcorpus, which presented values very similar to those of books, since the transcription of Critical Role sessions had not a scientific purpose. Although still low, the frequency of LLM-generated texts was higher than for books and human RPG.

\begin{figure}[htbp]
\centering
\subfigure[Length and Dept measures]{
  \includegraphics[width=0.48\textwidth]{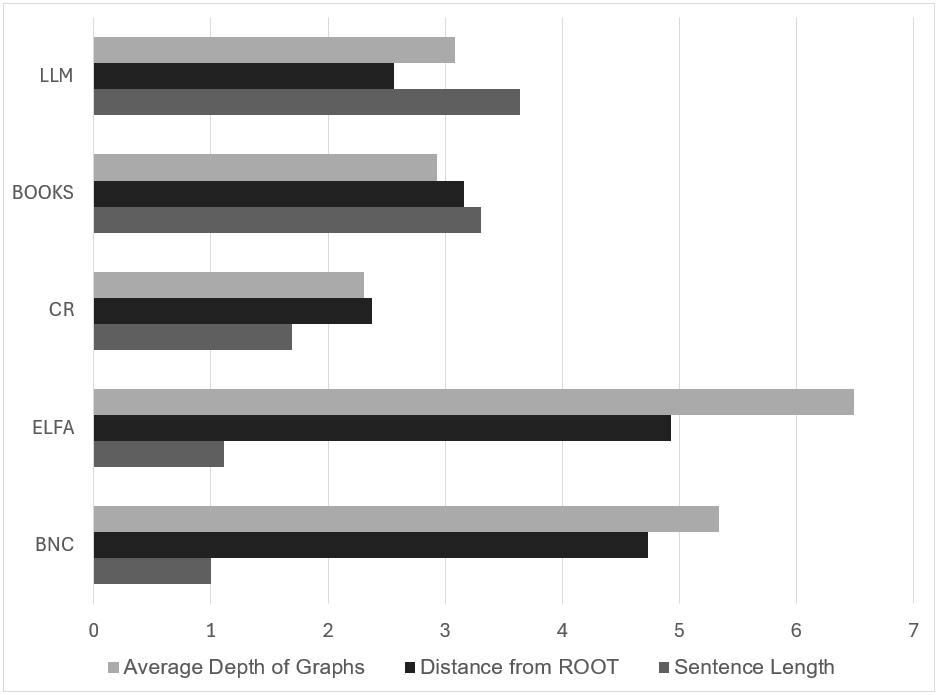}
  \label{fig:2a}
}
\hfill 
\subfigure[Clause and Nominal Modifier]{
  \includegraphics[width=0.46\textwidth]{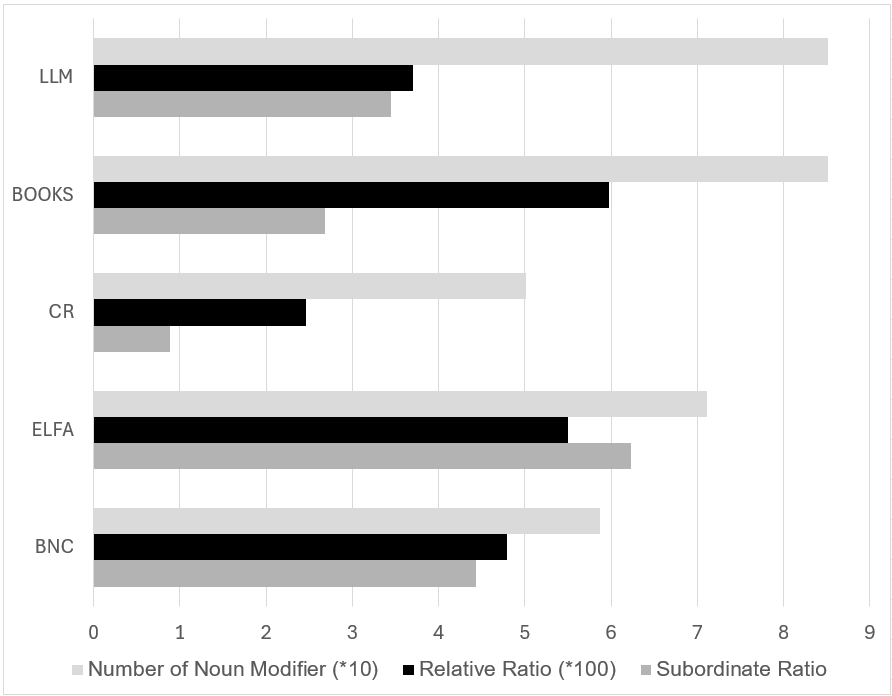}
  \label{fig:2b}
}
\caption{Measures of Syntactic Complexity}
\label{fig:2}
\end{figure}

Syntactic complexity measures often presented values that did not align with the expected features of spoken and written texts (see Figure \ref{fig:2}). While sentence length values were higher, as expected, for the \textit{book} and LLM RPG sub-corpora, the average depth of the dependency graph was much higher in \textit{conversation} and \textit{academic speech} (Figure \ref{fig:2a}). This high value of the distance from ROOT measure could be attributed to the nature of the recorded conversations, which include long oral presentations where sentences are not separated by punctuation. Figure \ref{fig:2b} also shows contradictory results: the number of subordinate clauses and relative clauses per sentence was surprisingly high in speech, except for the \textit{human RPG} sub-corpus, where the conversational nature of the D\&D sessions emerged with a very low number of subordinate and relative clauses. The number of noun modifiers reflected our expectations, with the highest values observed in books and LLM-generated texts.

Regarding verb tenses, the results align with the expected features of spoken and written texts. The use of the present tense is more pronounced in speech (higher than 0.50) compared to books (0.35). However, the LLM-generated texts exhibit a very high value (0.72). Books and LLM-generated texts reach the highest values for participles and gerunds, while the use of the past tense primarily characterizes books. First-person pronouns are much more frequent in speech.

Finally, we conducted a quantitative analysis of textual cohesion. The utilized cohesion metric confirmed the differences across the texts. The \textit{academic speech} corpus exhibited the highest cohesion value (0.212), followed by \textit{human RPG} (0.105). The \textit{book} sub-corpus achieved a cohesion score of 0.046, while \textit{conversation} obtained 0.026. The score obtained by LLM-generated texts was very low (0.007). Table \ref{tab:5} presents the obtained values.

\begin{table*}[htb]
\caption{Cohesion Measures}
\centering
\footnotesize
\begin{tabular}{l|lllll}
\toprule
\textbf{Values} & \textbf{BNC} & \textbf{ELFA} & \textbf{CR} & \textbf{BOOKS} & \textbf{LLM}\\
\midrule
Additive & 2.38 & 5.55 & 3.10 & 2.31 & 3.35\\
Causal & 4.66 & 7.76 & 6.55 & 4.78 & 3.71\\
Temporal & 3.50 & 2.69 & 2.81 & 5.58 & 3.14\\
Logical & 1.12 & 2.76 & 0.98 & 0.91 & 0.47\\
\midrule
Weighted Sum & 18.10 & 30.67 & 22.04 & 19.98 & 16.30\\
Standard Deviation & 1.08 & 2.07 & 1.38 & 1.15 & 1.02\\
\midrule
Cohesion Value & 0.026 & 0.212 & 0.105 & 0.046 & 0.007\\
\bottomrule
\end{tabular}
\label{tab:5}
\end{table*}

\subsection{Discussion}
The results of our experiment do not provide a definitive answer regarding the language register that LLMs adopt in their narrative strategies. From a lexical perspective, the D-value, a measure of lexical richness, highlights the differences between human and LLM-generated D\&D sessions. The D-values of the LLM sessions are much closer to those of books, likely reflecting the nature of the LLMs' training data, which primarily consists of written text. This influences the LLMs' ability to accurately reflect the style of a real RPG session. The analysis of lexical ranges supports this hypothesis: while we found similar LR1 values in Critical Role, books, and LLM-generated texts, the LLMs made more extensive use of terms included in the LR2 and LR3 dictionaries, indicating a greater reliance on less frequent and more academic vocabulary. Additionally, the concreteness value of Critical Role reflects its similarity to written texts, possibly due to the planned nature of the Dungeon Master's speech, while the results for the LLMs may indicate a lack of planning inherent in their generative process.  Finally, the most significant lexical finding is the high number of repetitions in the LLM-generated texts.

From a syntactic perspective, we first note that the Critical Role transcripts are much more similar to books than to the other speech categories. The syntactic data yielded two different and somewhat contradictory results: considering the sentence length and depth of the dependency graphs, we found that LLMs exhibit a pattern distinct from all other text categories. Specifically, they produce the longest sentences and the lowest distance from the root node. This indicates that the sentences produced by the LLMs are longer than those produced by humans, and their main element (the root) tends to be located near the center of the sentence. Regarding subordinate clauses, LLM-generated texts contain more subordinate clauses than books but a lower ratio of relative clauses, exceeding the values achieved by both in the Critical Role transcripts. Additionally, both LLMs and books exhibit a greater number of noun modifiers compared to the speech.

The analysis of verb tenses reveals a key difference between LLMs and the other texts. The LLMs almost exclusively used the present tense, occasionally replaced by participles, but rarely used the past tense. In Critical Role, the present tense is also dominant but interspersed with the past tense. This difference may stem from the LLMs' focus on the immediate actions and events of the narrative, while Critical Role includes numerous references to the characters' backstories and past events.

Finally, the analysis of textual cohesion presents some peculiarities, potentially due to the limited number of connectives included in the dictionary. For example, books exhibit a lower cohesion value compared to Critical Role or ELFA, possibly because more frequent connectives are often replaced by synonyms in books to avoid redundancy. Additionally, as Ong suggests, coordination is typically associated with orality. However, temporal connectives are most frequent in books, likely due to the greater importance of temporal connections between events in written narratives. In contrast, the LLMs obtained a very low cohesion value, primarily due to the lack of logical and causal connectives in their generated texts \cite{maisto2024harris}. Nevertheless, when calculating the correlation among the frequency of connectives in each text category, we found a strong correlation among all the spoken corpora but only a moderate correlation between books and LLM-generated texts.

As we commented for the qualitative analysis, the GM descriptions in LLM and human sessions are similar, while the PC dialogues in the human sessions reflect the spontaneous nature of spoken discourse.
 
The Critical Role (CR) transcripts differ from the other speech corpora in that the GM's language exhibits characteristics more akin to written language (Books), while the PC dialogues retain features typical of spoken language. For instance, in terms of lexical analysis, CR often falls between the speech corpora and Books, with exceptions observed in indicators such as deictic frequency and Lexical Range 3. Syntactically, CR displays average sentence length. However, the prevalence of short utterances in PC dialogues contributes to a lower proportion of subordinate and relative clauses, as well as a reduced Distance of the ROOT value. Cohesion analysis further confirms average values for CR.

Conversely, the LLM-generated dialogues present more irregular values across these linguistic measures. This irregularity may stem from the unique nature of the LLM-generated language, which blends features of both written and spoken discourse. While the dialogues are rich in descriptive language, complex sentences, and noun modifiers, they also exhibit extensive use of the present tense and a notably low degree of cohesion.

We decided to analyze the linguistic features of CR and LLM dialogues by dividing each session into two parts: the first containing only the DM dialogue lines (CR-DM and LLM-DM) and the second containing the PC dialogues (CR-PC and LLM-PC). As anticipated, the most significant differences emerged within the CR corpus. Table \ref{tab:comp} shows the features in which the differences are more evident. These metrics collectively indicate that PC dialogues are characterized by features of spoken language, whereas DM dialogues exhibit characteristics more akin to written text, such as that found in books.

\begin{table*}[htb]
\caption{Comparison between PCs and DM dialogues}
\centering
\footnotesize
\begin{tabular}{l|lll|lll}
\toprule
\textbf{Metrics} & \textbf{CR} & \textbf{CR-DM} & \textbf{CR-PC} & \textbf{LLM} & \textbf{LLM-DM} & \textbf{LLM-PC}\\
\midrule
D-value & 8.22 & 13.17 & 6.5 & 19.89 & 21.99 & 17.64 \\
Ratio Deictics/Articles &  0.16 & 0.27 & 0.12 & 0.41 & 0.37 & 0.40\\
Sentence Length & 1.69 & 2.35 & 1.46 & 3.63 & 3.69 & 3.31\\
Subordinate Ratio & 0.88 & 1.65 & 0.63 & 3.44 & 3.72 & 3.01\\
Relative Ratio (*100) & 2.46 & 4.75 & 1.73 & 3.70 & 4.96 & 3.09\\
Ratio Past Tense & 0.08 & 0.06 & 0.09 & 0.01 & 0.01 & 0.01\\
Ratio of First Person Pronouns & 0.16 & 0.06 & 0.20 & 0.12 & 0.07 & 0.12\\
Number of Noun Modifiers (*10) & 5.01 & 6.73 & 4.39 & 8.52 & 8.52 & 8.43\\
\bottomrule
\end{tabular}
\label{tab:comp}
\end{table*}

In contrast, the differences between LLM-DM and LLM-PC were primarily limited to the D-Value and the Ratio of First Person Pronouns, which also showed variations across the CR sub-corpora. This discrepancy is a direct result of the inherent nature of the dialogue: PCs are required to articulate their actions in the first person, whereas the DM frequently poses questions regarding their intended actions.

In conclusion, it is evident that, despite variations in prompt instructions, the language used by LLM-DM and LLM-PC is remarkably similar. Conversely, the human PC’s language exhibited numerous features of spoken discourse, often surpassing those observed in the two spoken corpora considered.

\section{Conclusion}
The results of our experiment offer insights into the narrative strategies of these LLMs with a similar architecture and training data to LLaMa3. Lexically, they demonstrate a rich vocabulary, with lexical diversity comparable to written texts, likely due to their extensive training data. They generate complex sentences with subordinate clauses and noun modifiers, approaching the syntactic complexity of the written text. However, analysis of concreteness and lexical repetition suggests that these LLMs generate text in a less planned manner than humans. Syntactically, these models produce longer sentences with a lower distance from the root node than human-generated text, indicating a unique sentence structure. They utilize more subordinate clauses than books, but fewer relative clauses, and more noun modifiers than speech. These findings suggest that they possess a distinct syntactic signature. A key difference between narratives generated by these models and humans lies in verb tenses. LLMs predominantly use the present tense, rarely employing the past tense. This contrasts with human narratives, which utilize a mix of present and past tenses to establish temporal context and character background. The analysis of textual cohesion reveals that LLMs exhibit low cohesion, with limited use of logical and causal connectives. This may contribute to less nuanced and persuasive narratives compared to human-generated stories. From this point of view, an LLM cannot replace the human GM in an RPG, but it can be effectively employed to simulate a non-player character upon precise instructions, introducing unpredictability to the narrative and ensuring greater variety to the story. Overall, these LLM models show potential for generating narratives, but their strategies differ significantly from human approaches \cite{lieto2021cognitive}. They appear better suited for generating formal written text, with limitations in replicating spoken language nuances and constructing temporally and textually complex narratives.

Further research is needed to understand the impact of training data on LLM narrative strategies, develop methods for improving textual cohesion and verb tense usage, and investigate reinforcement learning techniques for training LLMs to generate more human-like narratives, including affective analysis with systems like \cite{lieto2021commonsense}.

\bibliography{gdr}




\end{document}